%% file: main_arxiv.tex
\documentclass[sigconf]{acmart}

\usepackage{makecell}
\usepackage{bm}

\AtBeginDocument{%
  \providecommand\BibTeX{{%
    \normalfont B\kern-0.5em{\scshape i\kern-0.25em b}\kern-0.8em\TeX}}}

\setcopyright{acmcopyright}
\copyrightyear{2018}
\acmYear{2018}
\acmDOI{XXXXXXX.XXXXXXX}

\acmConference[Conference acronym 'XX]{Make sure to enter the correct
  conference title from your rights confirmation emai}{June 03--05,
  2018}{Woodstock, NY}
\acmPrice{15.00}
\acmISBN{978-1-4503-XXXX-X/18/06}

\acmSubmissionID{1633}

\newcommand{\qy}[1]{\textcolor{black}{#1}}

\begin{document}

\title{4DAC: Learning Attribute Compression for Dynamic Point Clouds}


\author{\mbox{Guangchi Fang$^1$ \hspace{10pt} Qingyong Hu$^2$ \hspace{10pt} Yiling Xu$^3$ \hspace{10pt}Yulan Guo$^{1\ast}$}}


\affiliation{$^1$ School of Electronics and Communication Engineering, Sun Yat-sen University\city{Shenzhen}\country{China}}
\affiliation{$^2$ Department of Computer Science, University of Oxford\city{Oxford}\country{United Kingdom}}
\affiliation{$^3$ Cooperative Medianet Innovation Center, Shanghai Jiaotong University\city{Shanghai}\country{China}}

\email{fanggch@mail2.sysu.edu.cn, qingyong.hu@cs.ox.ac.uk, yl.xu@sjtu.edu.cn}

\email{guoyulan@sysu.edu.cn}

\def\authors{Guangchi Fang, Qingyong Hu, Yiling Xu, Yulan Guo}
\renewcommand{\shortauthors}{Guangchi Fang, et al.}



\begin{abstract}


\qy{With the development of the 3D data acquisition facilities, the increasing scale of acquired 3D point clouds poses a challenge to the existing data compression techniques. Although promising performance has been achieved in static point cloud compression, it remains under-explored and challenging to leverage temporal correlations within a point cloud sequence for effective dynamic point cloud compression. In this paper, we study the attribute (\textit{e.g.}, color) compression of dynamic point clouds and present a learning-based framework, termed 4DAC. To reduce temporal redundancy within data, we first build the 3D motion estimation and motion compensation modules with deep neural networks. Then, the attribute residuals produced by the motion compensation component are encoded by the region adaptive hierarchical transform into residual coefficients. In addition, we also propose a deep conditional entropy model to estimate the probability distribution of the transformed coefficients, by incorporating temporal context from consecutive point clouds and the motion estimation/compensation modules. Finally, the data stream is losslessly entropy coded with the predicted distribution. Extensive experiments on several public datasets demonstrate the superior compression performance of the proposed approach.
}

\end{abstract}




\maketitle


\section{Introduction}
\label{sec:intro}

\input{chapters/010_intro}

\section{Related Work}
\label{sec:related_work}

\input{chapters/020_related}

\section{Methodology}
\label{sec:method}

\input{chapters/030_method}


\section{Experiments}
\label{Sec:Experiments}

\input{chapters/040_Experiments}

\section{Conclusion}

\input{chapters/050_conclusion}

\clearpage

\bibliographystyle{ACM-Reference-Format}
\bibliography{egbib}

\clearpage
\appendix

\section*{Appendix}

\input{chapters/070_Appendix}



\end{document}

%% file: chapters/010_intro.tex


\begin{figure}[t]
	\begin{center}
    	\includegraphics[width=1.0\linewidth]{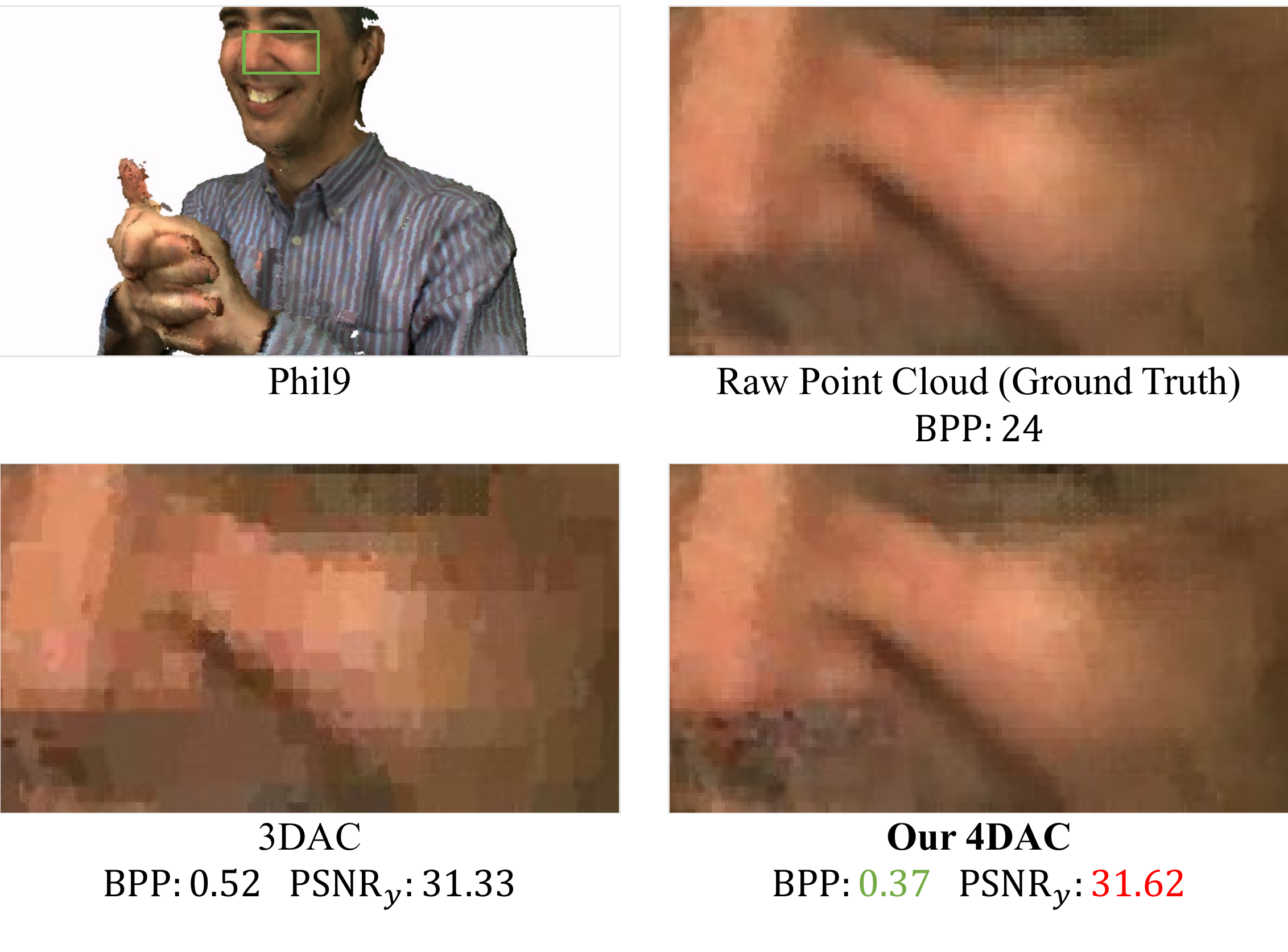}
	\end{center}
    \caption{Point cloud attribute compression results on the sequence “Phil” of the MVUB \cite{loop2016microsoft} dataset. The state-of-the-art learning-based static attribute compression method, 3DAC \cite{3dac}, is presented with our proposed 4DAC. Bits Per Point (BPP) and Peak Signal-to-Noise Ratio (PSNR) of the luminance component are also reported. Note that, raw RGB data of point cloud is usually stored as uint8 (\textit{i.e.,}  $8\times3=24$ BPP).
    }
	\label{fig:teaser_qual}
\end{figure}

\qy{As a common 3D representation, point cloud has been widely used in several real-world applications such as mixed reality \cite{mix_reality} and autonomous driving \cite{geiger2012we, hu2021learning} due to its compactness and simplicity. With the evolution of 3D laser scanning techniques and depth sensors, dynamic point clouds, which are composed of a sequence of static point clouds, have attracted increasing attention in recent years. Nevertheless, it remains a critical and challenging issue for the community on how to effectively compress unstructured point cloud streams. A handful of recent works, including MPEG standard \cite{schwarz2018emerging}, and several learning-based approaches in literature \cite{huang2020octsqueeze, biswas2020muscle, que2021voxelcontext, huang20193d, yan2019deep, quach2019learning, quach2020improved, gao2021point, wang2021multiscale} have shown encouraging performance on point cloud geometry compression (\textit{e.g.}, the 3D coordinates of points). However, the corresponding point cloud attributes (\textit{e.g.}, remission, color) have been largely overlooked by the community. Moreover, these attributes also take up a large amount of the overall bitstream in practice. Motivated by this, we mainly focus on attribute compression of dynamic point clouds in this paper.
}


\qy{Early works \cite{mekuria2016design, de2017motion, dorea2018block, dorea2019local} on dynamic point cloud attribute compression simply follow the common practice of video compression. Specifically, these methods usually follow a three-step pipeline, including 3D motion estimation and compensation, transform coding, and entropy coding. In particular, 3D motion estimation is the key component to reduce temporal redundancy within consecutive point clouds. Although encouraging effects has been achieved, the overall compression ratio is still limited by the hand-crafted motion estimation modules. Furthermore, other modules of the compression framework, such as entropy coding, are designed and optimized independently without considering the valuable temporal information.
}

\qy{Recently, a number of deep learning-based algorithms \cite{alexiou2020towards, sheng2021deep, isik2021lvac} are proposed for compressing point cloud attributes. In general, these methods build the transform coding or entropy coding module with the powerful deep neural networks, hence are likely to achieve comparable or even better performance than traditional hand-crafted compression algorithms. However, these methods mainly focus on the compression of static point clouds and do not consider the temporal correlations in consecutive point clouds. MuSCLE \cite{biswas2020muscle} is the only learning-based algorithm proposed to compress the intensity of LiDAR point cloud streams. Nevertheless, the compression performance is still limited by its naive framework, which does not have any motion estimation or transform coding blocks, and there is only spatial-temporal aggregation to extract temporal context for its entropy model.
}


\qy{In this paper, we present a learning-based attribute compression framework, termed 4DAC, for dynamic point clouds. Specifically, we propose two separate networks for 3D motion estimation and motion compensation. Region Adaptive Hierarchical Transform (RAHT) is adopted to encode the temporal attribute residuals produced by motion compensation to frequency-domain coefficients. Then, we further exploit temporal contexts to construct a deep conditional entropy model to estimate the probability distribution of the transformed coefficients. In particular, explicit and implicit temporal correlations hidden in consecutive point clouds and motion estimation/compensation blocks are explored for probability estimation. Finally, we feed the coefficients and their predicted distribution to an arithmetic coder to further compress the data stream losslessly. As shown in Fig. \ref{fig:teaser_qual}, by leveraging the useful temporal information, our method achieves better quantitative and qualitative compression performance than the static point cloud compression counterpart, 3DAC \cite{3dac}. Our key contributions are as follows:
}

\begin{itemize}
\setlength{\itemsep}{0pt}
\setlength{\parsep}{0pt}
\setlength{\parskip}{0pt}
    \item \qy{We propose an effective learning-based framework for dynamic point cloud compression, which is composed of a deep 3D motion estimation block, a deep 3D compensation block, and a deep entropy model.}
    \item \qy{We propose a temporal conditional entropy model to explore both explicit and implicit temporal correlation within data and motion estimation/compensation blocks.}
    \item \qy{Extensive experiments demonstrate the state-of-the-art compression performance achieved by our 4DAC on several point cloud benchmarks.}

\end{itemize}

%% file: chapters/020_related.tex
\subsection{Point Cloud Geometry Compression}

Point cloud geometry compression focuses on compressing the 3D coordinates of each point. Part of works \cite{meagher1982geometric, schwarz2018emerging, kammerl2012real, garcia2019geometry} rely on the octree structure to compress point cloud geometry. These works usually organize raw point cloud data as an octree and further entropy code the octants to a compact bitstream. To exploit temporal information within point cloud sequences, Kammerl et al. \cite{kammerl2012real} differentially encoded subsequent octree as XOR differences and Garcia et al. \cite{garcia2019geometry} introduced temporal context for entorpy coding. Recently, some learning-based compression methods \cite{huang2020octsqueeze, biswas2020muscle, que2021voxelcontext, fu2022octattention} also adopt the octree strcuture and construct their octree-based entropy models to reduce data redundancy. Inspired by deep image compression \cite{balle2018variational}, other learning-based approaches \cite{huang20193d, yan2019deep, quach2019learning, quach2020improved, gao2021point, wang2021multiscale} adopt 3D auto-encoders \cite{qi2017pointnet, brock2016generative, choy20194d} to achieve point cloud geometry compression. However, most of these methods are geometry-oriented and it is still a non-trivial task to extend them to attribute compression of dynamic point clouds.

\subsection{Point Cloud Attribute Compression}

Point cloud attribute compression aims at compressing the attributes (\textit{e.g.}, color) of points. Attribute compression algorithms can be roughly categorized as static or dynamic compression.

Static attribute compression methods \cite{zhang2014point, shao2017attribute, shao2018hybrid, de2016compression, souto2020predictive, pavez2021multi, pavez2020region, schwarz2018emerging, gu20193d} usually include two steps, transform coding of attributes, which captures data spatial redundancy in the transformed domain, and entropy coding of transformed coefficients, which losslessly compresses symbols into a compact bitstream. For example, Zhang et al. \cite{zhang2014point} encoded the voxliezed point cloud through graph transform and then entropy coded the transformed coefficients with a Laplacian probability distribution. Queiroz et al. \cite{de2016compression} proposed a 3D Haar wavelet transform, namely RAHT, and performed entropy coding with an adaptive run-length Golomb-Rice (RLGR) coder \cite{malvar2006adaptive}. Other recent works applied deep learning techniques for attribute compression. Alexiou et al. \cite{alexiou2020towards} and Sheng et al. \cite{sheng2021deep} built the transform coding and entropy coding blocks using a 3D auto-encoder with 3D convolution \cite{achlioptas2018learning} and point convolution \cite{qi2017pointnet++}, respectively. Fang et al. \cite{3dac} presented a learning-based attribute compression framework with RAHT for transform coding and a RAHT-based deep entropy model for entropy coding. In general, these approaches are designed for static point clouds and do not leverage the temporal information within dynamic point clouds.


In contrast, dynamic attribute compression algorithms \cite{mekuria2016design, de2017motion, thanou2015graph, thanou2016graph, biswas2020muscle, souto2020predictive} usually consider temporal correlations within point clouds. Compared with the aforementioned static compression framework, most prior works on dynamic attribute compression additionally include 3D motion estimation. Mekuria et al. \cite{mekuria2016design} split the raw point cloud as blocks and adopted rigid transformations, ICP, for motion estimation. Queiroz and Chou \cite{de2017motion} also used the block representation and transferred motion estimation as the task of block-based feature matching. Thanou et al. \cite{thanou2015graph, thanou2016graph} represented point cloud sequences as a set of graphs and conducted feature extraction and matching on graphs. These approaches rely on hand-crafted motion estimation blocks to reduce temporal redundancy and transmitted attribute residuals through the original static compression framework. In other words, blocks including entropy coding are not optimized with the temporal information. Thus, this simple combination does not make full use of temporal correlations and limits the compression performance.


\begin{figure*}[t]
	\centering
	\includegraphics[width=1.0\linewidth]{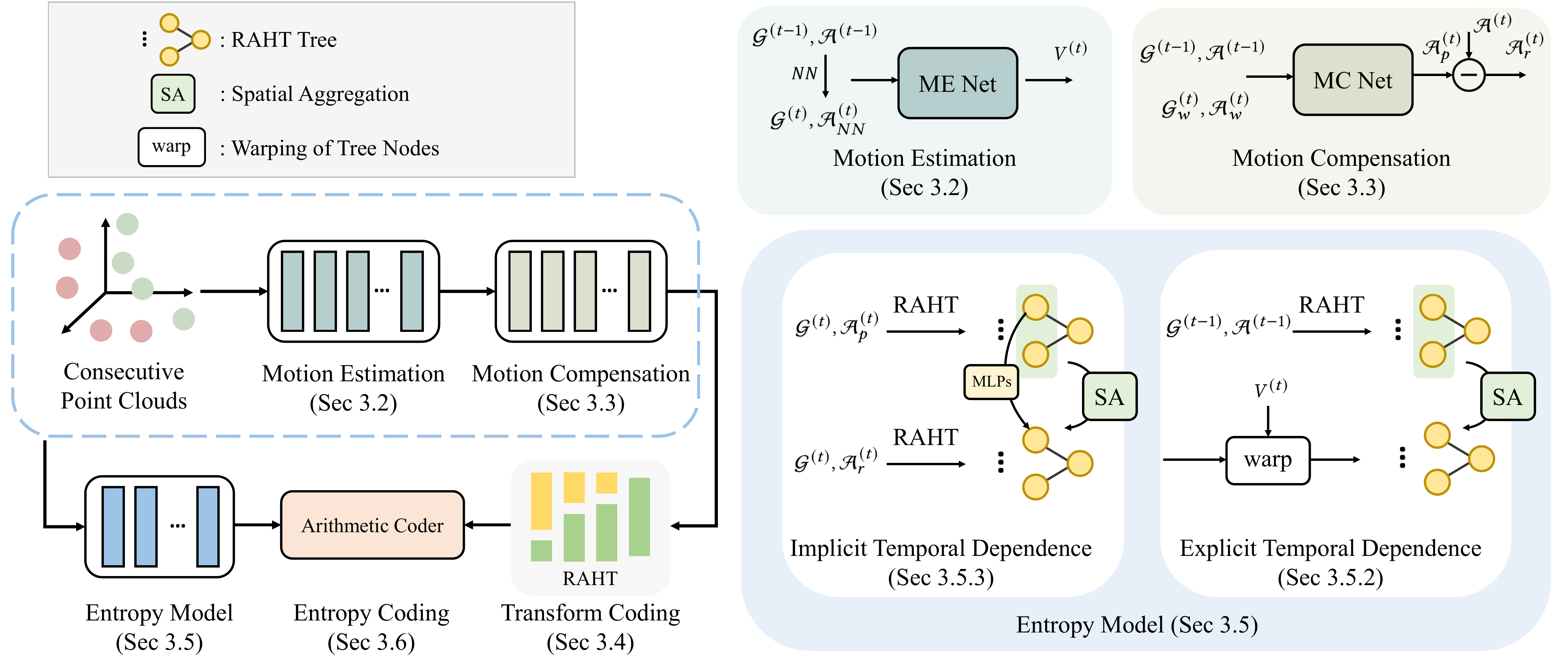}
	
    \caption{The network architecture of our dynamic point cloud attribute compression method, 4DAC.}
	\label{fig:method_overview}
\end{figure*}

\subsection{Deep Video Compression}

In the past decades, several video compression standards \cite{schwarz2007overview, sullivan2012overview, bross2021developments} have been proposed. Most methods adopted hand-crafted motion estimation/compensation and residual coding to reduce the temporal and spatial redundancies. Recently, the promising deep learning techniques bring a new research direction, deep video compression, to the data compression community and a number of learning-based approaches \cite{wu2018video, rippel2019learned, agustsson2020scale, lu2019dvc, hu2021fvc} have been proposed. For example, Wu et al. \cite{wu2018video} combined an image compression model and a interpolation model for video compression. Lu et al. \cite{lu2019dvc} proposed a learning-based video compression framework following the traditional hybrid video compression pipeline. In this work, they utilized an emerging 2D flow estimation network for motion estimation and used the warped frame for motion compensation. Hu et al. \cite{hu2021fvc} further performed operations of the compression framework in the latent feature space for better compression. Inspire by these approaches, we aim to construct 3D motion estimation/compensation modules for dynamic point cloud compression.

\subsection{Motion Estimation}

Motion estimation is important for both video and point cloud compression. 2D motion estimation is also referred as optical flow estimation \cite{dosovitskiy2015flownet, ilg2017flownet, sun2018pwc, teed2020raft}, which aims at finding pixel-wise motions between consecutive images. Recently, 2D motion estimation has been extended to 3D domain with point convolution \cite{liu2019flownet3d, kittenplon2021flowstep3d} and 3D convolution \cite{gojcic2021weakly, li2021sctn}. These works show the potential of 3D motion estimation and compensation for dynamic point cloud compression.

%% file: chapters/030_method.tex
\subsection{Overview}

Given a point cloud sequence $\left\{\mathcal{P}^{(1)}, \ldots, \mathcal{P}^{(t)}\right\}$, the point cloud geometry $\left\{\mathcal{G}^{(1)}, \ldots, \mathcal{G}^{(t)}\right\}$ and the part of attributes $\left\{\mathcal{A}^{(1)}, \ldots, \mathcal{A}^{(t-1)}\right\}$ are assumed to have been transmitted, and we focus on the compression of the current point cloud attributes $\mathcal{A}^{(t)}$. In Fig. \ref{fig:method_overview}, we present our 4DAC, a learning-based attribute compression framework for dynamic point clouds. Without loss of generality, our framework can be extended to point clouds with other attributes, such as plenoptic point clouds \cite{sandri2018compression}.

As shown in Fig. \ref{fig:method_overview}, we first estimate 3D motion vectors from consecutive point clouds, and then perform motion compensation with the estimated 3D motion to obtain temporal attribute residuals. Next, we encode the residuals to transformed coefficients through region adaptive hierarchical transform (RAHT) \cite{de2016compression}. In addition, we propose a deep temporal conditional entropy model to estimate distribution of transformed coefficients. Specifically, we explore implicit and explicit temporal correlations hidden in consecutive point clouds and motion estimation/compensation blocks for our deep entropy model. Finally, the coefficients and their estimated distributions are sent to an arithmetic coder to produce a compact bitstream in the entropy coding stage.

\subsection{Motion Estimation}\label{sec:Motion Estimation}

In this step, we estimate motion vectors $V^{(t)}$ from two consecutive point clouds, $\mathcal{P}^{(t-1)}$ and $\mathcal{P}^{(t)}$. The predicted motion vectors $V^{(t)}$ will be used for point cloud warping in Secs. \ref{sec:Motion Compensation} and \ref{sec:Explicit Temporal Dependence}. In order to preserve local geometry and attribute patterns of the reference point cloud $\mathcal{P}^{(t-1)}$ in the aforementioned sections, we set the motion vectors $V^{(t)}$ pointing from the current source point cloud $\mathcal{P}^{(t)}$ to the previous reference point cloud $\mathcal{P}^{(t-1)}$ and thus we can warp $\mathcal{P}^{(t)}$ instead of $\mathcal{P}^{(t-1)}$. It is also worth noting that, a relative reasonable flow estimation result can be obtained without the target attributes $\mathcal{A}^{(t)}$. Benefiting from this result, we can simply predict 3D motion information from point cloud geometry and do not need to transmit motion vectors. In particular, we feed $\left\{\mathcal{G}^{(t-1)}, \mathcal{A}^{(t-1)}\right\}$ and $\left\{\mathcal{G}^{(t)}, \mathcal{A}^{(t)}_{NN}\right\}$ to the flow estimation network, FlowNet3D \cite{liu2019flownet3d}, where $\mathcal{A}^{(t)}_{NN}$ is the nearest neighbor prediction from $\left\{\mathcal{G}^{(t-1)}, \mathcal{A}^{(t-1)}\right\}$ to $\mathcal{G}^{(t)}$. More specifically, for each point in $\mathcal{G}^{(t)}$, we find its nearest neighbor in $\mathcal{G}^{(t-1)}$ and use the attributes of the nearest neighbor as the nearest neighbor prediction attributes.

\subsection{Motion Compensation}\label{sec:Motion Compensation}

In order to capture temporal redundancy, the motion compensation block obtains attribute residuals $\mathcal{A}^{(t)}_{r}$ from consecutive point clouds, $\mathcal{P}^{(t-1)}$ and $\mathcal{P}^{(t)}$, and the estimated motion vectors $V^{(t)}$. As mentioned in Sec. \ref{sec:Motion Estimation}, we first warp the target point cloud geometry $\mathcal{G}^{(t)}$ with the motion vectors $V^{(t)}$, and the warped geometry $\mathcal{G}^{(t)}_{w}$ is obtained as $\mathcal{G}^{(t)}_{w}=\mathcal{G}^{(t)}+V^{(t)}$. Then, we construct a 3D attribute embedding network based on FlowNet3D \cite{liu2019flownet3d} to estimate the target attributes $\mathcal{A}^{(t)}$. Specifically, we simply set the nearest neighbor prediction $\mathcal{A}^{(t)}_{w}$ from the $\left\{\mathcal{G}^{(t-1)}, \mathcal{A}^{(t-1)}\right\}$ to $\mathcal{G}^{(t)}_{w}$ as the initial attribute prediction, and refine $\mathcal{A}^{(t)}_{w}$ to the final prediction $\mathcal{A}^{(t)}_{p}$ through the proposed network. Finally, the temporal attribute residuals $\mathcal{A}^{(t)}_{r}$ can be obtained (\textit{i.e.,}  $\mathcal{A}^{(t)}_{r}=\mathcal{A}^{(t)}-\mathcal{A}^{(t)}_{p}$). The network architecture details of the motion estimation and compensation networks are provided in appendix.

\subsection{Transform Coding}\label{sec:Transform Coding}

We further encode the temporal attribute residuals $\mathcal{A}^{(t)}_{r}$ into transformed coefficients $\mathcal{C}^{(t)}_{r}$ to reduce spatial redundancy. For simplicity, we use Region Adaptive Hierarchical Transform (RAHT) \cite{de2016compression} for transform coding. In short, RAHT is an effective 3D variant of the wavelet transform algorithm, which hierarchically decomposes attributes to DC and high-frequency coefficients. The DC coefficient is transmitted directly and the high-frequency coefficients will be further entropy coded in Sec. \ref{sec:Entropy Coding}. This step is similar to most traditional attribute compression methods \cite{de2016compression, souto2020predictive, pavez2021multi, pavez2020region}.

\begin{figure}[t]
	\centering
	\includegraphics[width=1.0\linewidth]{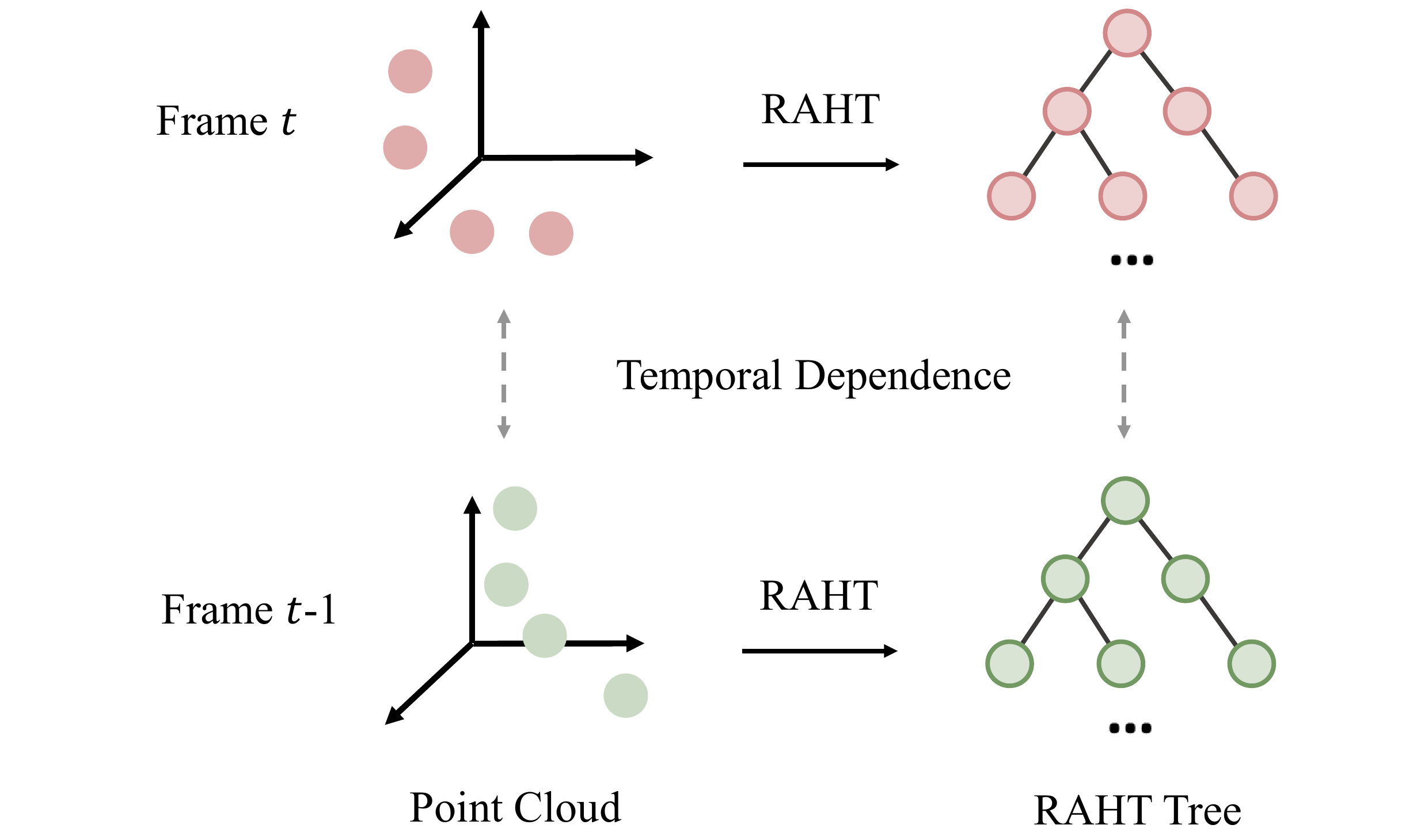}
    \caption{A simple illustration of RAHT and RAHT tree.}
	\label{fig:method_raht}
\end{figure}

According to the hierarchical transform steps, we can construct RAHT trees following \cite{3dac}. As shown in Fig. \ref{fig:method_raht}, for each point cloud in the sequence, we are able to build a RAHT tree, whose high-frequency tree nodes contain the transformed coefficients. Intuitively, there is a temporal dependence between two consecutive point clouds. Furthermore, it is also possible to extend this temporal dependence to two consecutive RAHT trees. In 3DAC \cite{3dac}, transform coding context and inter-channel correlation are explored via the tree structure for static point cloud attribute compression. In this paper, we further build temporal correlations within dynamic point clouds through the RAHT trees in Secs. \ref{sec:Explicit Temporal Dependence} and \ref{sec:Implicit Temporal Dependence}. The details about RAHT and RAHT tree structure can be found in \cite{de2016compression, 3dac} and the appendix.

\subsection{Entropy Model}\label{sec:Entropy Model}

The objective of the entropy model is to estimate the actual distribution of the input symbols. For dynamic point cloud attribute compression, one of the main difficulties is how to model the temporal correlations between two point clouds without direct correspondence. In this section, we propose our temporal conditional entropy model based on the RAHT tree structure. We first formulate the distribution of the transformed attribute residual coefficients $q(\mathcal{C}^{(t)}_{r})$ with temporal conditions, and then adopt neural networks to model this formulation.

\subsubsection{Formulation}

Given the geometry sequence $\left\{\mathcal{G}^{(1)}, \ldots, \mathcal{G}^{(t)}\right\}$ and attribute sequence $\left\{\mathcal{A}^{(1)}, \ldots, \mathcal{A}^{(t-1)}\right\}$, we need to compress and transmit the current attributes $\mathcal{A}^{(t)}$. After motion estimation, motion compensation and transform coding (Secs. \ref{sec:Motion Estimation}, \ref{sec:Motion Compensation} and \ref{sec:Transform Coding}), the task has been converted to compress the transformed attribute residual coefficients $\mathcal{C}^{(t)}_{r}=\left\{\bm{c}^{(t)(1)}_{r}, \ldots, \bm{c}^{(t)(m)}_{r}\right\}$. We factorize probability distribution $q(\mathcal{C}^{(t)}_{r})$ as follows:

\begin{equation}
q(\mathcal{C}^{(t)}_{r})=\prod_{i} q\left(\bm{c}^{(t)(i)}_{r} \mid {\mathcal{G}^{(1)}, \ldots, \mathcal{G}^{(t)}}, {\mathcal{A}^{(1)}, \ldots, \mathcal{A}^{(t-1)}}\right).
\end{equation}
For simplicity, we assume the current point cloud $\mathcal{P}^{(t)}$ only depends on the previous transmitted point cloud $\mathcal{P}^{(t-1)}$. Thus, the formulation can be simplified as:

\begin{equation}
q(\mathcal{C}^{(t)}_{r})=\prod_{i} q\left(\bm{c}^{(t)(i)}_{r} \mid {\mathcal{G}^{(t)}, \mathcal{G}^{(t-1)}}, {\mathcal{A}^{(t-1)}}\right).
\end{equation}
Note that, it is also possible to buffer the transmitted point cloud sequence to capture more extensive temporal information. 

Given the current point cloud geometry $\mathcal{G}^{(t)}$ and the previous point cloud geometry and attributes, $\mathcal{G}^{(t-1)}$ and $\mathcal{A}^{(t-1)}$, we obtained intermediate information including the motion vectors $V^{(t)}$ and the attribute prediction from motion compensation $\mathcal{A}^{(t)}_{p}$ in Secs. \ref{sec:Motion Estimation} and \ref{sec:Motion Compensation}. Next, we will exploit the aforementioned context information and conclude two types of temporal context (explicit temporal dependence and implicit temporal dependence) to model the distribution $q\left(\bm{c}^{(t)(i)}_{r} \mid {\mathcal{G}^{(t)}, \mathcal{G}^{(t-1)}}, {\mathcal{A}^{(t-1)}}\right)$ with neural networks.

\begin{figure}[t]
	\centering
	\includegraphics[width=1.0\linewidth]{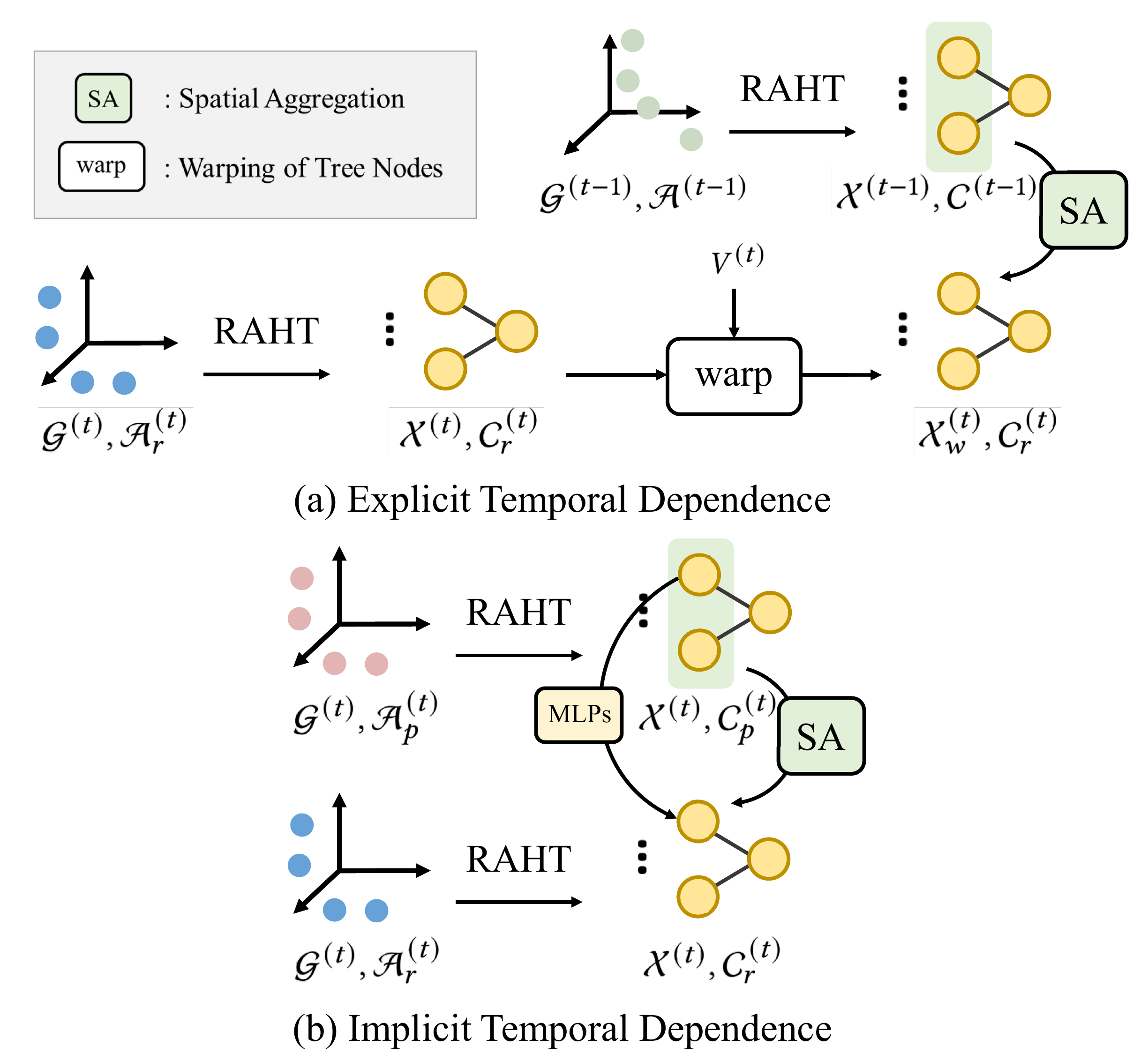}
    \caption{Our proposed temporal context. (a) Explicit temporal dependence between two consecutive point clouds, (b) Implicit temporal dependence between target attributes and prediction from motion compensation.}
	\label{fig:temporal_dependence}
\end{figure}

\subsubsection{Explicit Temporal Dependence}\label{sec:Explicit Temporal Dependence}

Our explicit temporal dependence module explores the direct temporal correlations between two consecutive point clouds. One straightforward approach provided by MuSCLE \cite{biswas2020muscle} is to extract spatial-temporal context feature by performing point convolution \cite{wang2018deep} on point cloud geometry. However, this naive temporal context aggregation method is highly based on their framework, which directly transmits point cloud attributes without transform coding. According to \cite{3dac}, this simple framework may limit the compression performance. Thus, in our proposed framework, we conduct temporal context feature extraction on two consecutive RAHT trees.

As shown in Fig. \ref{fig:temporal_dependence}(a), we utilize the RAHT tree structure to represent dynamic point clouds. In particular, we first transform the previous point cloud $\{\mathcal{G}^{(t-1)}, \mathcal{A}^{(t-1)}\}$ to a RAHT tree ${\{\mathcal{X}^{(t-1)}, \mathcal{C}^{(t-1)}}\}$, where $\mathcal{X}$ and $\mathcal{C}$ denote positions and transformed coefficients of tree nodes, respectively. Similarly, attribute residual with the current geometry $\{\mathcal{G}^{(t)}, \mathcal{A}^{(t)}_{r}\}$ is converted to $\{\mathcal{X}^{(t)}, \mathcal{C}^{(t)}_{r}\}$. In order to minimize the spatial difference between two RAHT trees $\mathcal{X}^{(t)}$ and $\mathcal{X}^{(t-1)}$, we further warp $\mathcal{X}^{(t)}$ to $\mathcal{X}^{(t-1)}$ with the motion vectors of tree nodes. Note that, the motion vectors of tree nodes can be simply obtained as the mean value of $V^{(t)}$ in the corresponding subspace.

Then, we employ spatial aggregation to bridge two consecutive RAHT trees. Here, we have the reference RAHT tree ${\{\mathcal{X}^{(t-1)}, \mathcal{C}^{(t-1)}}\}$ and the warped tree ${\{\mathcal{X}^{(t)}_{w},\mathcal{C}^{(t)}_{r}}\}$. In this step, we propose our explicit temporal context feature by exploring temporal correlations between the transmitted coefficients $\mathcal{C}^{(t-1)}$ and the transformed coefficients $\mathcal{C}^{(t)}_{r}$. It is worth noting that $\mathcal{C}^{(t)}_{r}$ is highly related to its temporal-spatial neighbors in $\mathcal{C}^{(t-1)}$. Thus, we aggregate those neighbors through PointNet++ \cite{qi2017pointnet++} for temporal feature aggregation. Specifically, at each depth level of RAHT trees, the explicit temporal context feature $\mathbf{E}_{i}$ can be obtained as follows:

\begin{equation}
\mathbf{E}_{i}=\underset{j \in N(i)}{\operatorname{MAX}}\left\{M L P s\left(\bm{c}^{(t-1)(j)}, \bm{x}^{(t-1)(j)}-\bm{x}_{w}^{(t)(i)}\right)\right\},
\end{equation}
where $\bm{x}^{(t-1)(j)}$ and $\bm{c}^{(t-1)(j)}$ is the tree node $j$ of ${\{\mathcal{X}^{(t-1)},\mathcal{C}^{(t-1)}}\}$, and $\bm{x}^{(t)(i)}$ is tree node $i$ of $\mathcal{X}^{(t)}_{w}$. $N(i)$ is the k-nearest temporal-spatial neighbors of $\bm{x}^{(t)(i)}_{w}$ in the same depth level of $\mathcal{X}^{(t-1)}$. $MAX$ is the element-wise
max pooling and $MLPs$ denotes Multi-Layer Perceptrons.

\subsubsection{Implicit Temporal Dependence}\label{sec:Implicit Temporal Dependence}

Our implicit temporal dependence module incorporates intermediate temporal information hidden in 3D motion processing parts. Considering this type of temporal information is extracted from dynamic points indirectly, we denote it as implicit temporal context in contrast to the aforementioned explicit temporal context. Here, we make use of attribute prediction $\mathcal{A}^{(t)}_{p}$ from our motion compensation network to estimate $\mathcal{A}^{(t)}_{r}$. As shown in Fig. \ref{fig:temporal_dependence}(b), point clouds $\{\mathcal{G}^{(t)}, \mathcal{A}^{(t)}_{p}\}$ and $\{\mathcal{G}^{(t)}, \mathcal{A}^{(t)}_{r}\}$ are firstly transformed to RAHT trees ${\{\mathcal{X}^{(t)},\mathcal{C}^{(t)}_{p}}\}$ and ${\{\mathcal{X}^{(t)},\mathcal{C}^{(t)}_{r}}\}$. Note that, these two RAHT trees share the same spatial structure $\mathcal{X}^{(t)}$. In other words, the attribute prediction and the residual coefficients, $\bm{c}_{p}^{(t)(i)}$ and $\bm{c}_{r}^{(t)(i)}$, share the same tree node $i$. Thus, we extract latent node feature $\mathbf{N}_{i}$ from $\bm{c}_{p}^{(t)(i)}$ with MLPs:

\begin{equation}
\mathbf{N}_{i}=M L P s\left(\bm{c}_{p}^{(t)(i)}\right).
\end{equation}
We can also extract spatial feature $\mathbf{S}_{i}$ within the RAHT tree through point convolution:


\begin{equation}
\mathbf{S}_{i}=\underset{j \in N(i)}{\operatorname{MAX}}\left\{M L P s\left(\bm{c}^{(t)(j)}_{p}, \bm{x}^{(t)(j)}-\bm{x}^{(t)(i)}\right)\right\}.
\end{equation}
Finally, we fuse $\mathbf{N}_{i}$ and $\mathbf{S}_{i}$ to obtain our implicit temporal context embedding $\mathbf{I}_{i}$:
\begin{equation}
\mathbf{I}_{i}=M L P s\left(\mathbf{N}_{i}, \mathbf{S}_{i}\right).
\end{equation}

\subsubsection{Probability Estimation}

We further aggregate the context features and adopt a probability model \cite{balle2018variational} to model the distribution of residual coefficients. In particular, we feed two temporal context features $\mathbf{E}_{i}$ and $\mathbf{I}_{i}$ into MLPs and incorporate the generated feature into a fully factorized density model \cite{balle2018variational}, which model the distribution $q\left(\bm{c}^{(t)(i)}_{r} \mid {\mathcal{G}^{(t)}, \mathcal{G}^{(t-1)}}, {\mathcal{A}^{(t-1)}}\right)$ as $q\left(\bm{c}^{(t)(i)}_{r} \mid {\mathbf{E}_{i}, \mathbf{I}_{i}}\right)$.

\subsection{Entropy Coding}\label{sec:Entropy Coding}

At the entropy coding stage, we feed the coefficients $\mathcal{C}^{(t)}_{r}$ and their corresponding distribution $q\left(\bm{c}^{(t)(i)}_{r} \mid {\mathbf{E}_{i}, \mathbf{I}_{i}}\right)$ to an arithmetic coder to produce the final attribute bitstream.

\subsection{Learning}

During training, we optimize our motion estimation network, motion compensation network and deep entropy model jointly. We adopt L2 loss for both motion estimation and compensation:
\begin{equation}\label{eq:mv_loss}
\mathrm{L}_{\mathrm{ME}}=\frac{1}{m} \sum_{i}^{m}\left\{\left(v_{p}^{(t)(i)}-v_{g t}^{(t)(i)}\right)^{2}\right\},
\end{equation}
where $v_{g t}^{(t)(i)}$ and $v_{p}^{(t)(i)}$ are the ground truth motion vector and the predicted motion vector of point $i$ in the point cloud $\mathcal{P}^{(t)}$, and
\begin{equation}
\mathrm{L}_{\mathrm{MC}}=\frac{1}{m} \sum_{i}^{m}\left\{\left(\bm{a}_{p}^{(t)(i)}-\bm{a}_{g t}^{(t)(i)}\right)^{2}\right\},
\end{equation}
where $\bm{a}_{g t}^{(t)(i)}$ and $\bm{a}_{p}^{(t)(i)}$ are the ground truth attributes and the predicted attributes of the point $i$. For our deep entropy model, we use the cross-entropy loss:
\begin{equation}
\mathrm{L}_{\mathrm{CE}}=-\sum_{i} \log q(\bm{c}_{r}^{(t)(i)} \mid \mathbf{E}_i, \mathbf{I}_i).
\end{equation}
Thus, the total loss $\mathrm{L}_{\mathrm{total}}$ is composed of the motion estimation loss $\mathrm{L}_{\mathrm{ME}}$, the motion compensation loss $\mathrm{L}_{\mathrm{MC}}$ and the cross entropy loss $\mathrm{L}_{\mathrm{CE}}$:
\begin{equation}
\mathrm{L}_{\mathrm{total}}=\mathrm{L}_{\mathrm{CE}}+\lambda_{\text {ME }}\mathrm{L}_{\mathrm{ME}}+\lambda_{\text {MC }}\mathrm{L}_{\mathrm{MC}},
\end{equation}
where $\lambda_{\text {ME }}$ and $\lambda_{\text {MC }}$ are set to 1.0 empirically.

%% file: chapters/040_Experiments.tex
We evaluate our 4DAC first quantitatively and then qualitatively on several dynamic point cloud datasets. Then, we include two downstream tasks to further demonstrate the effectiveness of our proposed method. We also conduct extensive ablation studies on each component of our framework.

\subsection{Experimental Setup}

\noindent\textbf{(1) Datasets.} For evaluation of our method, we use three benchmark datasets as follows:
\begin{itemize}
\setlength{\itemsep}{1pt}
\setlength{\parsep}{1pt}
\setlength{\parskip}{1pt}
  \item \textbf{FlyingThings3D \cite{mayer2016large, liu2019flownet3d}}. This is a dynamic point cloud dataset generated by scenes with moving objects in ShapeNet \cite{chang2015shapenet}. This dataset contains 22k pairs of colored point clouds with ground truth motion vectors. Following the official training/testing split \cite{liu2019flownet3d}, we use 20k of them for training and 2k for test. 
  
  \item \textbf{MVUB \cite{loop2016microsoft}}. This is a dynamic point cloud dataset introduced by \cite{de2016compression}. There are five subject sequences in the dataset, named as "Andrew", "David", "Ricardo", "Phil" and "Sarah". These colored point cloud sequences are generated by RGBD cameras with a voxelization algorithm \cite{loop2013real}. Each sequence contains 200 to 300 point clouds. We use "Andrew", "David" and "Ricardo" for training, and "Phil" and "Sarah" for test.

  \item \textbf{8iVFB \cite{d20178i}}. This is a MPEG/JPEG point cloud compression dataset containing four human point cloud sequences, including "Soldier", "Longdress", "Loot" and "Redandblack". Each point cloud sequence is captured by RGB cameras and contains 300 dense point clouds. We use "Soldier" and "Longdress" for training, and others for test.
  
\end{itemize}

\noindent\textbf{(2) Evaluation Metrics.} Following \cite{de2016compression}, we report the Peak Signal-to-Noise Ratio of the luminance component ($\rm {PSNR}_y$) and Bits Per Point ($\rm BPP$) to evaluate the reconstruction quality and the compression ratio, respectively.

\noindent\textbf{(3) Baselines.} We compare our 4DAC with several baselines:
\begin{itemize}
\setlength{\itemsep}{1pt}
\setlength{\parsep}{1pt}
\setlength{\parskip}{1pt}
    \item \textbf{RAHT \cite{de2016compression}}. This is a baseline method for both static and dynamic point cloud attribute compression. Following the original implementation in \cite{de2016compression}, we use RAHT for transform coding and the run-length Golomb-Rice coder \cite{malvar2006adaptive} for entropy coding.
    
    \item \textbf{NN-RAHT}. This is a baseline method for dynamic point cloud attribute compression. Based on the aforementioned RAHT, we additionally include nearest neighbor prediction. More specifically, we assume the motion vectors of point clouds equal to zero and adopt nearest neighbor for motion estimation and motion compensation. The attribute residual is further encoded by RAHT and a run-length Golomb-Rice coder.
    
    \item \textbf{G-PCC \cite{schwarz2018emerging}}. This is a MPEG standard point cloud compression software \footnote{We use version 13.0 of G-PCC: https://github.com/MPEGGroup/mpeg-pcc-tmc13}. G-PCC supports three transform coding methods and several other options. For a fair comparison, we choose RAHT without inter-depth prediction for transform coding. 
    
    \item \textbf{3DAC \cite{3dac}}. This is a state-of-the-art learning-based attribute compression method focusing on static point clouds. In 3DAC \cite{3dac}, RAHT is adopted for transform coding and a tree-structured deep entropy model is proposed for entropy coding.
    
\end{itemize}
Note that, we also conduct experiments on other learning-based attribute compression methods on our benchmarks. Considering that the results of these methods are beyond the common quality comparison range (\textit{i.e.}, low $\rm {PSNR}_y$ with high $\rm BPP$), we separately present them in appendix for better illustration.

\begin{figure*}[t]
	\centering
	\includegraphics[width=1.0\linewidth]{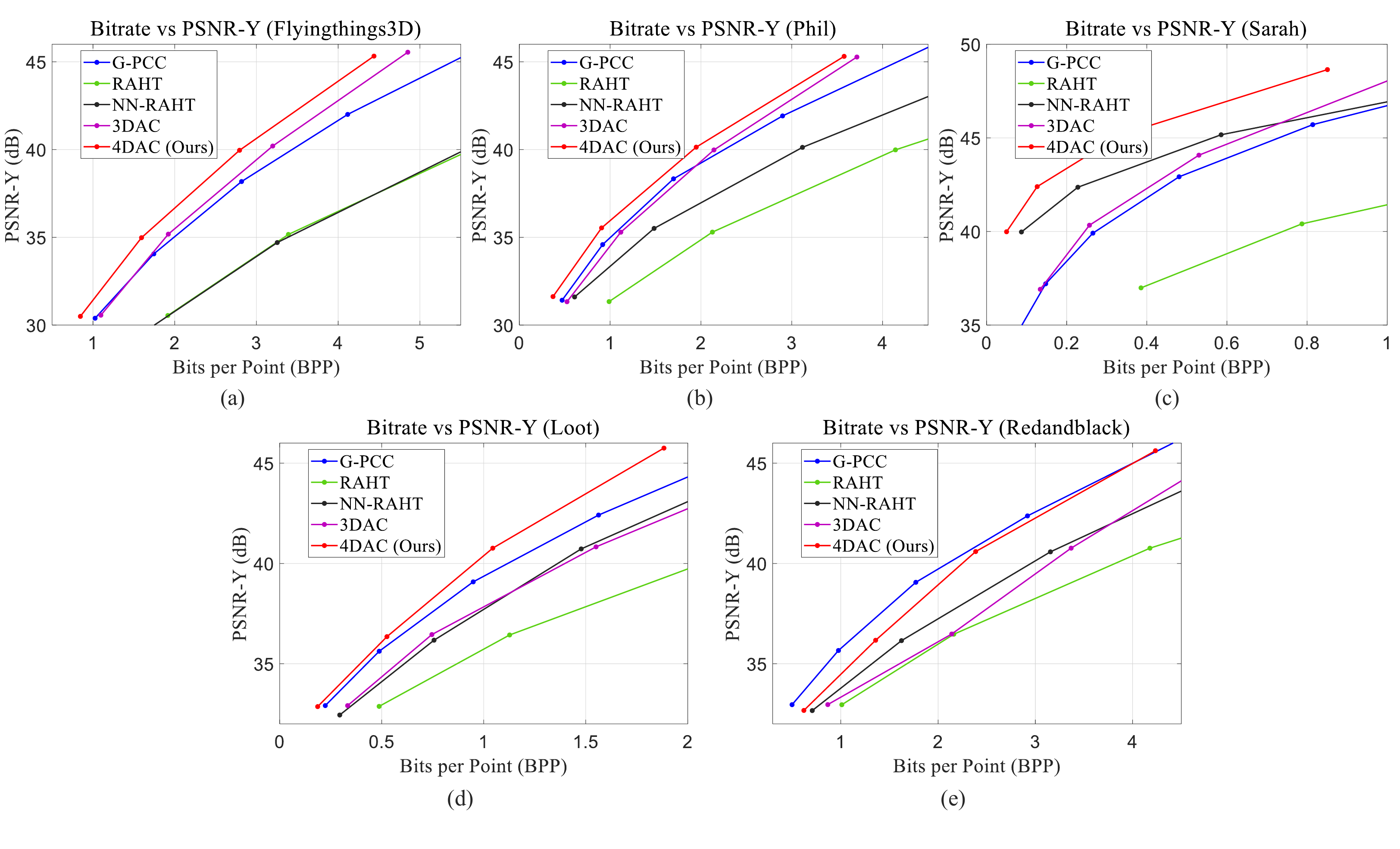}
	
    
    \caption{Quantitative results on FlyingThings3D (a), MVUB "Phil"(b), MVUB "Sarah"(c), 8iVFB "Loot" (d) and 8iVFB "Redandblack" (e).}    
    
	\label{fig:paper_quant}
\end{figure*}

\begin{figure}[t]
	\begin{center}
    	\includegraphics[width=1.0\linewidth]{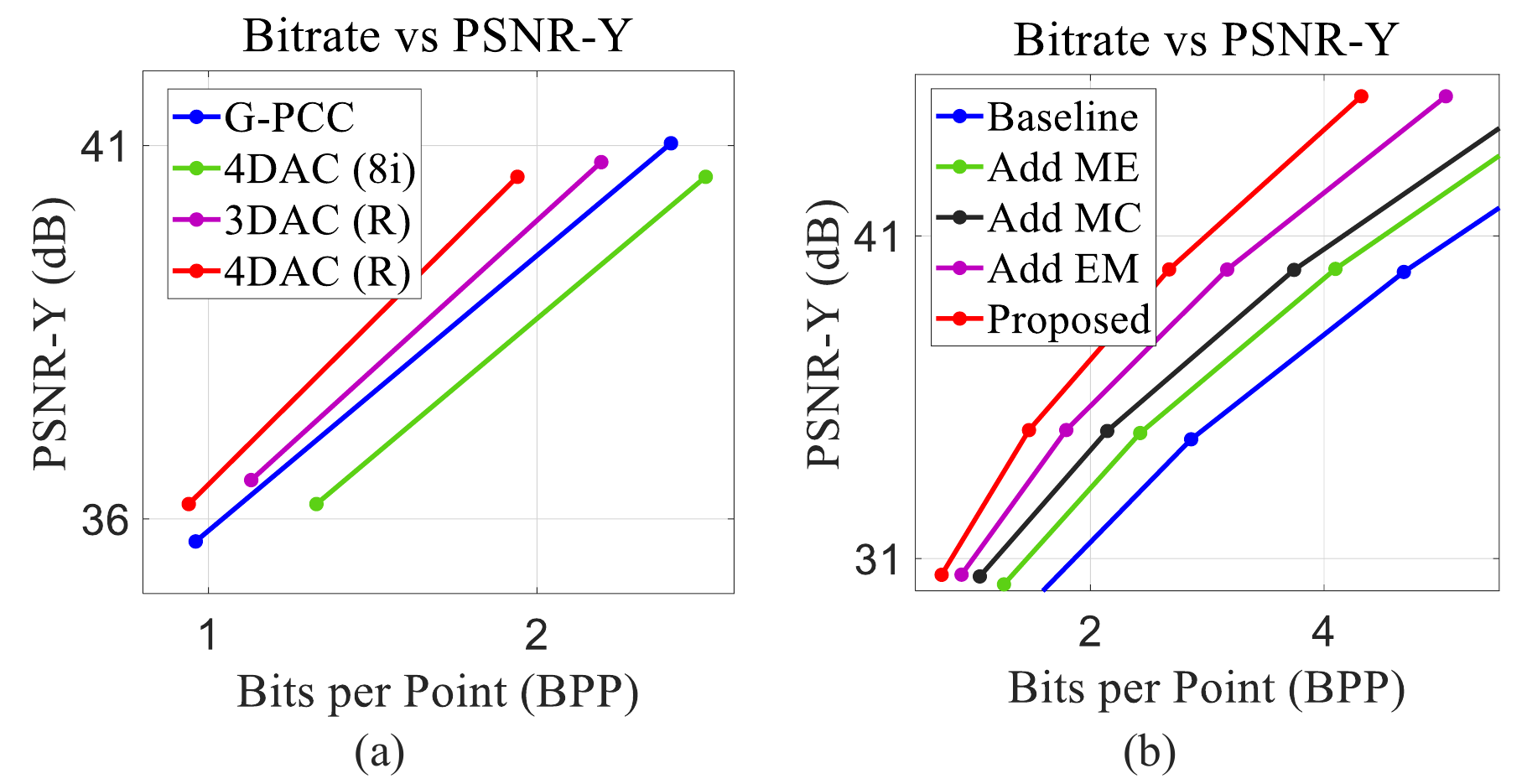}
	\end{center}
    \caption{
    Experiments about the training dataset and our proposed modules.
    (a) Effect of the training dataset. We train models on the training set of 8iVFB and "Redandblack", respectively, and evaluate them on the test set of "Redandblack".
    (b) Ablation study on FlyingThings3D. We add motion estimation, motion compensation and entropy model one by one, and report the corresponding attribute compression performance.
    }
	\label{fig:exp_ablation}
\end{figure}

\noindent\textbf{(4) Implementation Details.} In the common case of point cloud compression, point cloud geometry and attributes are usually compressed and transmitted separately. To simulate this real condition in our experiments, we use a 9-level octree to voxelize the raw point cloud data and assume the point cloud geometry has been transmitted independently. For all datasets, we convert point cloud colors from the RGB color space to the YUV color space following the default setting of G-PCC \cite{schwarz2018emerging}. Because there is no ground truth motion vector in the MVUB and 8iVFB datasets, we adopt a pseudo motion vector as the ground truth $v_{g t}^{(t)}$ in Eq. \ref{eq:mv_loss}. More specifically, for a point in the current point cloud frame, we search its nearest neighbor in the previous point cloud through spatial and color space (\textit{i.e.}, xyzrgb) and use the spatial displacement between two points as the pseudo motion vector. Note that, more advanced methods, such as self-supervised scene flow estimation, can be utilized, which is beyond the scope of this paper. Considering that our 4DAC and the previous 3DAC \cite{3dac} share the similar transform coding and entropy coding blocks, we further include context features of 3DAC with our temporal context.


\subsection{Evaluation on Public Datasets}

Figure \ref{fig:paper_quant} shows the quantitative attribute compression results on FlyingThings3D, MVUB (sequences "Phil" and "Sarah") and 8iVFB (sequences "Loot" and "Redandblack"). We can see that NN-RAHT outperforms RAHT on MVUB and 8iVFB because these datasets contain point clouds with relative small motion. However, on FlyingThings3D, nearest neighbor motion estimation has no positive effect due to the relative large motion of dynamic point cloud pairs. In contrast, our 4DAC consistently outperforms its static counterpart, 3DAC, by a large margin on all benchmarks, which shows the effectiveness of our dynamic compression framework. Overall, our method achieves competitive compression performance compared with other baselines on most datasets except "Redandblack". The main reason is that the color distribution of "Redandblack" is different from our training sequences (\textit{i.e.}, "Soldier" and "Longdress"). 

To validate this assumption, we adopt a train-val-test split of [50\%, 20\%, 30\%] on the sequence "Redandblack". Note that, point clouds in the sequence is not shuffled, and thus we can simulate a use case that we are compressing subsequent point cloud frames with a transmitted point cloud sequence. The result is shown in Fig. \ref{fig:exp_ablation}(a), 3DAC (R) and 4DAC (R) denote 3DAC and 4DAC trained on the train split of "Redandblack", and 4DAC (8i) is 4DAC trained on "Soldier" and "Longdress" of 8iVFB. Both 3DAC (R) and 4DAC (R) outperform G-PCC and 4DAC (8i), which shows the effect of the training dataset. One simple but practical solution is to enlarge the scale of the training dataset to minimize the domain gap between the training and test splits.

We also provide the qualitative results on "Loot" with the corresponding quantitative results in Fig. \ref{fig:exp_qual}. The results on other datasets are provided in appendix. It is obvious that our method outperforms other attribute compression methods, which is consistent to the quantitative results.

\begin{figure*}[t]
	\centering
	\includegraphics[width=1.0\linewidth]{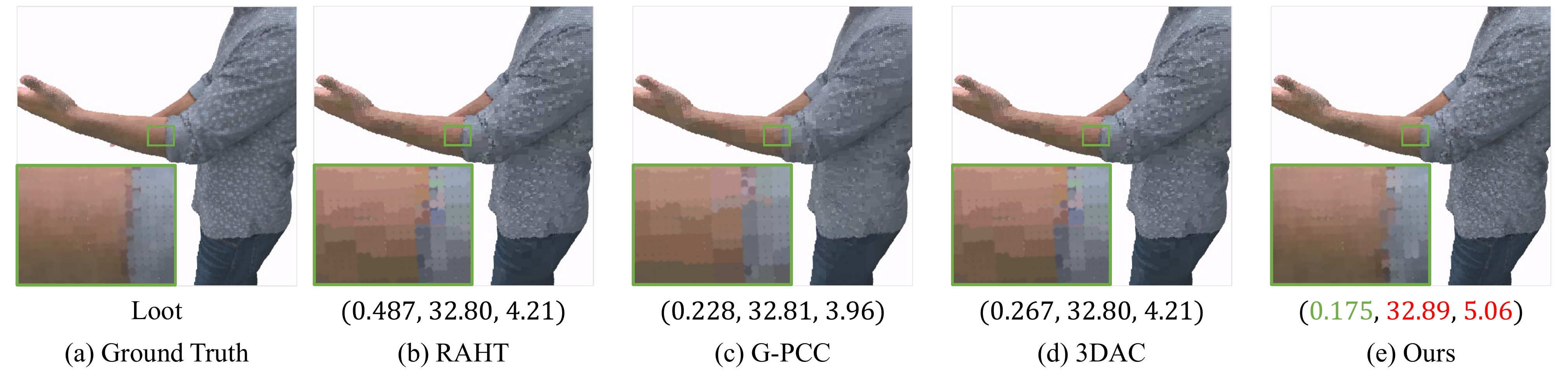}
	
    \caption{Qualitative results of our 4DAC and other baselines. The quantitative results ($\rm BPP$, $\rm {PSNR}_y$, $\rm GraphSIM$) are also shown under the figure. our 4DAC achieves the best reconstruction quailty ($\rm {PSNR}_y$ and $\rm GraphSIM$) with the lowest bitrate.}
	\label{fig:exp_qual}
\end{figure*}

\begin{figure}[t]
	\begin{center}
    	\includegraphics[width=1.0\linewidth]{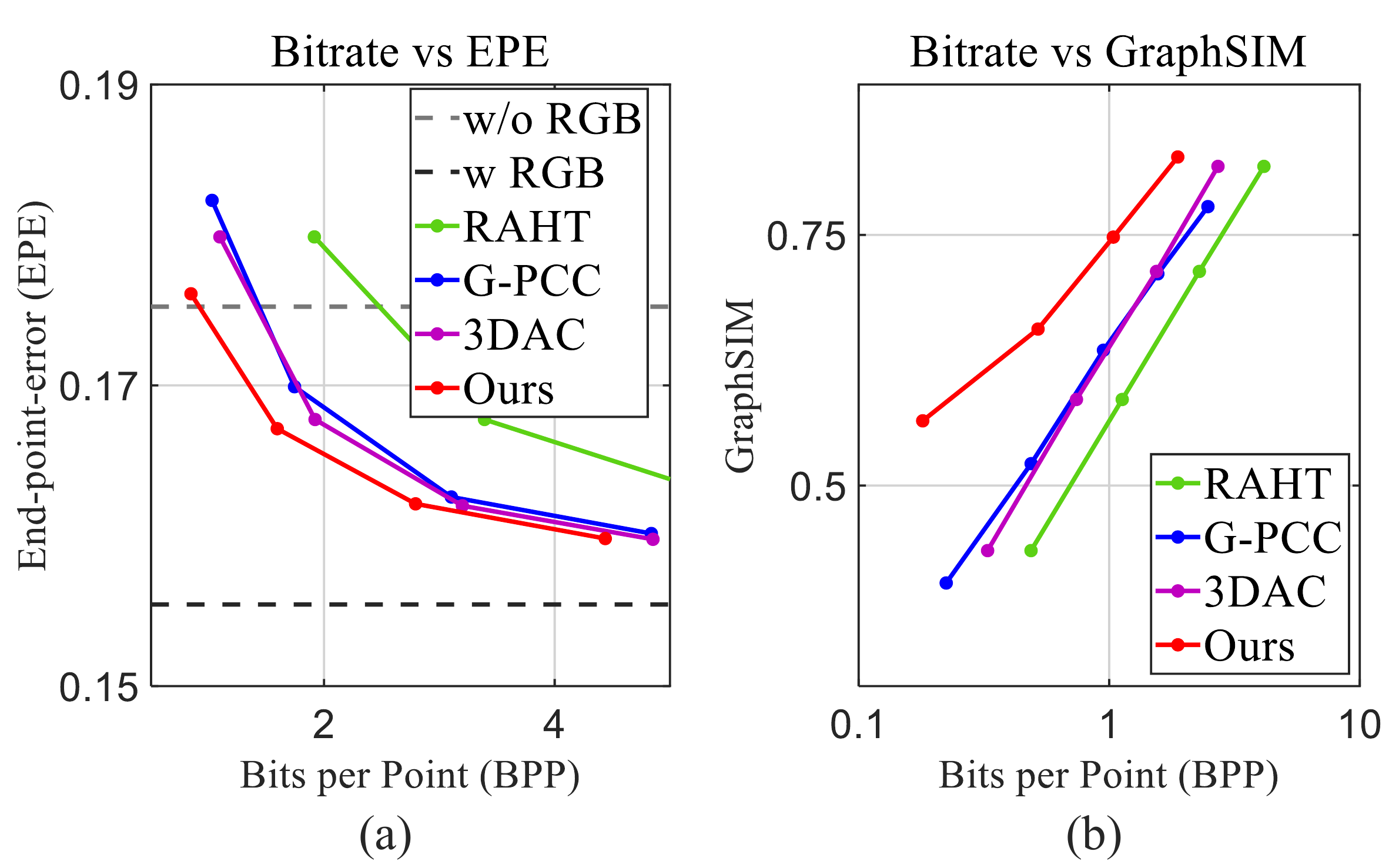}
	\end{center}
    \caption{The experiment results on two downstream tasks. (a) Scene flow estimation on FlyingThings3D. (b) Quality asessment on "Loot".
    }
	\label{fig:exp_downstream}
\end{figure}

\subsection{Evaluation on Downstream Tasks}

In order to validate the compression performance on downstream tasks, we conduct experiments on scene flow estimation and quality assessment for machine and human perception, respectively.

\textbf{Scene Flow Estimation.} Figure \ref{fig:exp_downstream}(a) shows the downstream performance on FlyingThings3D with FlowNet3D \cite{liu2019flownet3d}. Following \cite{3dac}, we train the network on the raw point clouds and evaluate it with the compressed target point cloud. End-point-error (EPE) at different $\rm BPP$s is used as the evaluation metric, and the results of raw point clouds with and without attributes are denoted as `w RGB' and `w/o RGB', respectively. It is obvious that our 4DAC outperforms 3DAC and G-PCC, which demonstrates the superior compression performance of our method on this downstream task.

\textbf{Quality Assessment.} We adopt the point cloud quality assessment task to show the effectiveness of our method on human perception. In particular, we use GraphSIM \cite{yang2020inferring} to evaluate the point cloud attribute reconstruction quality. As an attribute-sensitive metric, a higher GraphSIM score means better reconstruction quality. As shown in Fig. \ref{fig:exp_downstream}(b), with the help of temporal information, our approach outperforms other baselines by a large margin. This experimental result further demonstrates the effectiveness of our method.

\subsection{Ablation Study}

In this section, we conduct ablation studies on each component of our dynamic attribute compression framework to show their effect on the compression performance. As shown in Fig. \ref{fig:exp_ablation}(b), we include several results by adding our proposed components progressively. In particular, we construct a basic baseline method with RHAT as transform coding and a fully factorized density model \cite{balle2018variational} as entropy model, and denote it as `Baseline'. Wu further construct `Add ME', `Add MC', `Add EM' by adding motion estimation, motion compensation, deep entropy model with temporal context. We additionally incorporate 3DAC context and denote the final framework as `proposed'. Experiments are conducted on FlyingThings3D with different $\rm BPP$s. The performance shown in Fig. \ref{fig:exp_ablation}(b) demonstrates that our framework can achieve a much lower bitrate with the similar reconstruction quality with the help of our proposed components. Besides, the result also indicates that our 4DAC can perform well with the previous state-of-the-art compression algorithm, 3DAC \cite{3dac}. It shows the possibility for our framework to incorporate with more advanced algorithms for a better compression performance.

We also report the ablation studies on our temporal context features for our deep entropy model in Table \ref{tab:ablation_input}. Specifically, we also conduct experiments on FlyingThings3D and set the uniform quantization parameter as 10 for the same reconstruction quality. We progressively incorporate explicit temporal context (\textbf{E}), implicit temporal context (\textbf{I}) and 3DAC context (\textbf{3DAC}), and it can be seen that the bitrate is steadily reduced with incorporation of more context features, which shows the effectiveness of our proposed explicit and implicit temporal dependence modules.

\input{Tables/ablation_em}

%% file: Tables/ablation_em.tex
\begin{table}[htbp]
\small 
  \centering
  \resizebox{0.21\textwidth}{!}{
    \begin{tabular}{ccc|c}
    \Xhline{2.0\arrayrulewidth}
    E & I & 3DAC & Birate \\
    \hline
             &       &       & 3.74   \\
$\checkmark$ &       &       & 3.53  \\
$\checkmark$ & $\checkmark$ &          & 3.17   \\
$\checkmark$ & $\checkmark$ & $\checkmark$ & 2.67  \\
          \Xhline{2.0\arrayrulewidth}
    \end{tabular}%
    }

\caption{Ablation study on context features for entropy model. 
E, I and 3DAC denote explicit temporal context, implicit temporal context and 3DAC context \cite{3dac}, respectively.}   
  \label{tab:ablation_input}%
\end{table}%

%% file: chapters/050_conclusion.tex
We have presented a dynamic point cloud attribute compression framework. This framework includes a motion estimation block, a motion compensation block and a temporal conditional entropy model to leverage temporal information within data. Besides, our deep entropy model exploits explicit and implicit temporal dependence to model the probability distribution. Furthermore, we show the competitive performance of our method on several datasets, and the results on downstream tasks show that our algorithm is capable for both machine and human perception tasks.

%% file: chapters/070_Appendix.tex
\section{Additional Method Details}

\subsection{Motion Estimation and Compensation}

\begin{figure}[t]
	\centering
	\includegraphics[width=1.0\linewidth]{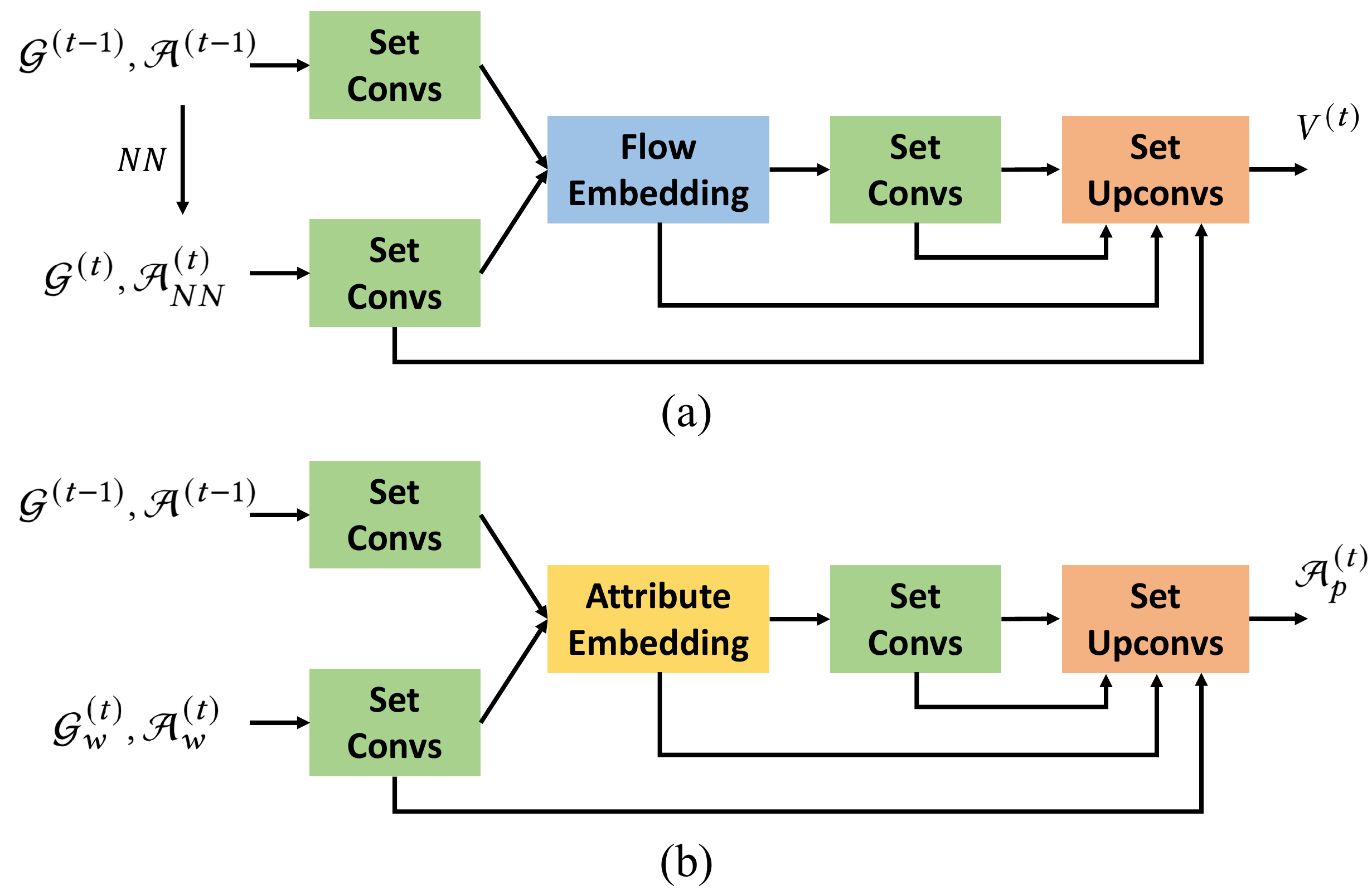}
    \caption{The proposed Motion Estimation Net (a) and Motion Compensation Net (b).
    }
	\label{fig:memc}
\end{figure}

The motion estimation network and the motion compensation network are shown in Fig. \ref{fig:memc}. We construct our motion estimation and motion compensation nets with modules (\emph{Set Conv}, \emph{Flow Embedding} and \emph{Set Upconv}) proposed by FlowNet3D \cite{liu2019flownet3d}. Because the motion compensation net extracts point cloud attribute information rather than motion information, we denote the attribute feature fusion module (the yellow block in Fig. \ref{fig:memc}(b)) as \emph{Attribute Embedding} instead of \emph{Flow Embedding} although they share the similar architecture.

\subsection{Transform Coding}\label{sec:Transform Coding appendix}

In this paper, we use Region Adaptive Hierarchical Transform (RAHT) \cite{de2016compression} for transform coding. In the following parts, we first introduce RAHT, and then show how to construct a RAHT tree.

\subsubsection{Region Adaptive Hierarchical Transform}\label{sec:Region Adaptive Hierarchical Transform appendix}

RAHT transforms point cloud attributes to frequency-domain coefficients based on 3D Haar wavelet transform. Specifically, the algorithm takes a voxelized point cloud as input, and convert attributes to low- and high-frequency coefficients along three dimensions repeatedly. Different from the common 2D wavelet transform, RAHT is able to process empty spaces through transmitting low-frequency coefficients to the next depth level directly. Besides, the wavelet basis of RAHT is adaptive to the number of points in the corresponding region. More specifically, two low-frequency neighbors are decomposed to low- and high-frequency coefficients as follows:
\begin{equation}
\left[\begin{array}{c}
l_{d+1, x, y, z} \\
h_{d+1, x, y, z}
\end{array}\right]=\frac{1}{\sqrt{w_{1}+w_{2}}}\left[\begin{array}{cc}
\sqrt{w_{1}} & \sqrt{w_{2}} \\
-\sqrt{w_{2}} & \sqrt{w_{1}}
\end{array}\right]\left[\begin{array}{c}
l_{d, 2 x, y, z} \\
l_{d, 2 x+1, y, z}
\end{array}\right],
\end{equation}
where $l_{d, 2 x, y, z}$ and $l_{d, 2 x+1, y, z}$ are two low-frequency neighbors along the $x$ dimension, and $w_{1}$ and $w_{2}$ are their weights (\textit{i.e.}, the number of points in the corresponding subspace). $l_{d+1, x, y, z}$ and $h_{d+1, x, y, z}$ are the decomposed low- and high-frequency coefficients, respectively.

\begin{figure}[t]
	\centering
	\includegraphics[width=1.0\linewidth]{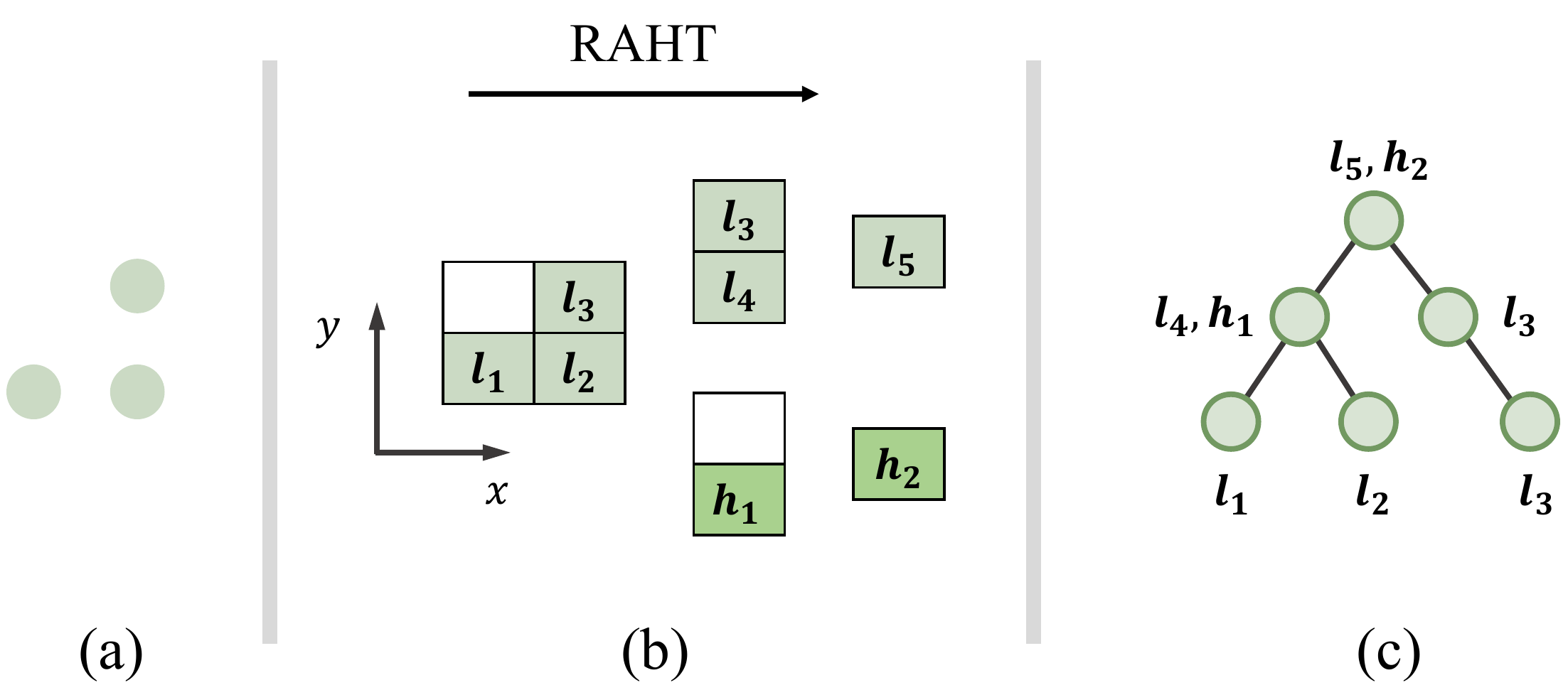}
    \caption{A 2D example of RAHT. (a) Point cloud. (b) The transform steps of RAHT. (C) The corresponding RAHT tree. 
    }
	\label{fig:appendix_raht}
\end{figure}

\subsubsection{RAHT Tree}\label{sec:RAHT Tree appendix}

We revisit the transform process of RAHT and the construction of a RAHT tree through a 2D toy example in Fig. \ref{fig:appendix_raht}. The input point cloud is first voxlized and then transformed. Low- and high-frequency coefficients are denoted as $l_{i}$ and $h_{i}$, respectively. During encoding, transform is conducted along the $x$ axis and then the $y$ axis repeatedly until all voxelized points are merged. At the first depth level, along the $x$ axis, point without neighbor (\textit{i.e.}, $l_{3}$), is transmitted to the next depth level and those with a neighbor (\textit{i.e.}, $l_{1}$ and $l_{2}$) are decomposed to low- and high-frequency coefficients (\textit{i.e.}, $l_{4}$ and $h_{1}$). In this way, we transform low-frequency coefficients depth by depth until generating the DC coefficient $l_{5}$. After forward transform, we compress the DC coefficient and all high-frequency coefficients, and thus, during decoding, we can recover all low-frequency coefficients. Specifically, decoding is start from the root level. We can reconvert the DC coefficient $l_{5}$ and high-frequency coefficient $h_{2}$ to low-frequency coefficients of the previous depth (\textit{i.e.}, $l_{3}$ and $l_{4}$). In this way, all low-frequency coefficients can be regenerated depth by depth.

Here, we further construct the RAHT tree of the aforementioned toy example. Following the hierarchical transform processes, the RAHT tree is constructed as Fig. \ref{fig:appendix_raht}(c). Low-frequency coefficients of the first depth level (\textit{i.e.}, point cloud attributes), $l_{1}$, $l_{2}$ and $l_{3}$, are set as leaf nodes. These tree nodes are transmitted or merged according to the repeated transform steps. Specifically, tree nodes of $l_{1}$ and $l_{2}$ are merged to that of $l_{4}$ and $h_{1}$, and tree nodes of $l_{4}$ and $l_{3}$ are merged to that of $l_{5}$ and $h_{2}$.



\section{Additional Implementation Details}

We pretrain the motion estimation and compensation networks first and then train these two networks with our deep entropy model jointly. For faster convergence of the motion estimation net, we initialize the estimated motion vectors $V^{(t)}$ as zeros. In specific, we multiply the output vectors of the motion estimation net with learnable tensors which are initialized as zeros, and use the result as the estimated motion vectors $V^{(t)}$. For our motion compensation net, we initialize the predicted attributes $\mathcal{A}^{(t)}_{p}$ as the input attributes $\mathcal{A}^{(t)}_{w}$. In specific, we multiply the output vectors of the motion compensation net with learnable tensors (which are also initialized as zeros), and add them with the input attributes. The result is regarded as the the predicted attributes. For Explicit Temporal Dependence, we aggregate temporal features from 3 nearset neighbors through 3-layer MLPs (16, 32 and 8 dimensional hidden features). For Implicit Temporal Dependence, we extract latent node feature $\mathbf{N}_{i}$ with 3-layer MLPs (16, 32 and 8) and spatial feature $\mathbf{S}_{i}$ from 3 nearset neighbors with 3-layer MLPs (16, 32 and 8), then we fuse the node feature and spatial feature to our implicit temporal context embedding $\mathbf{I}_{i}$ with a MLP (8). Finally, the explicit and implicit context features are concatenated with the 3DAC context feature for our deep entropy model.

\section{Additional Experiments}



\subsection{Comparison with Learning-based Methods}

\begin{figure*}[t]
	\centering
	\includegraphics[width=0.95\linewidth]{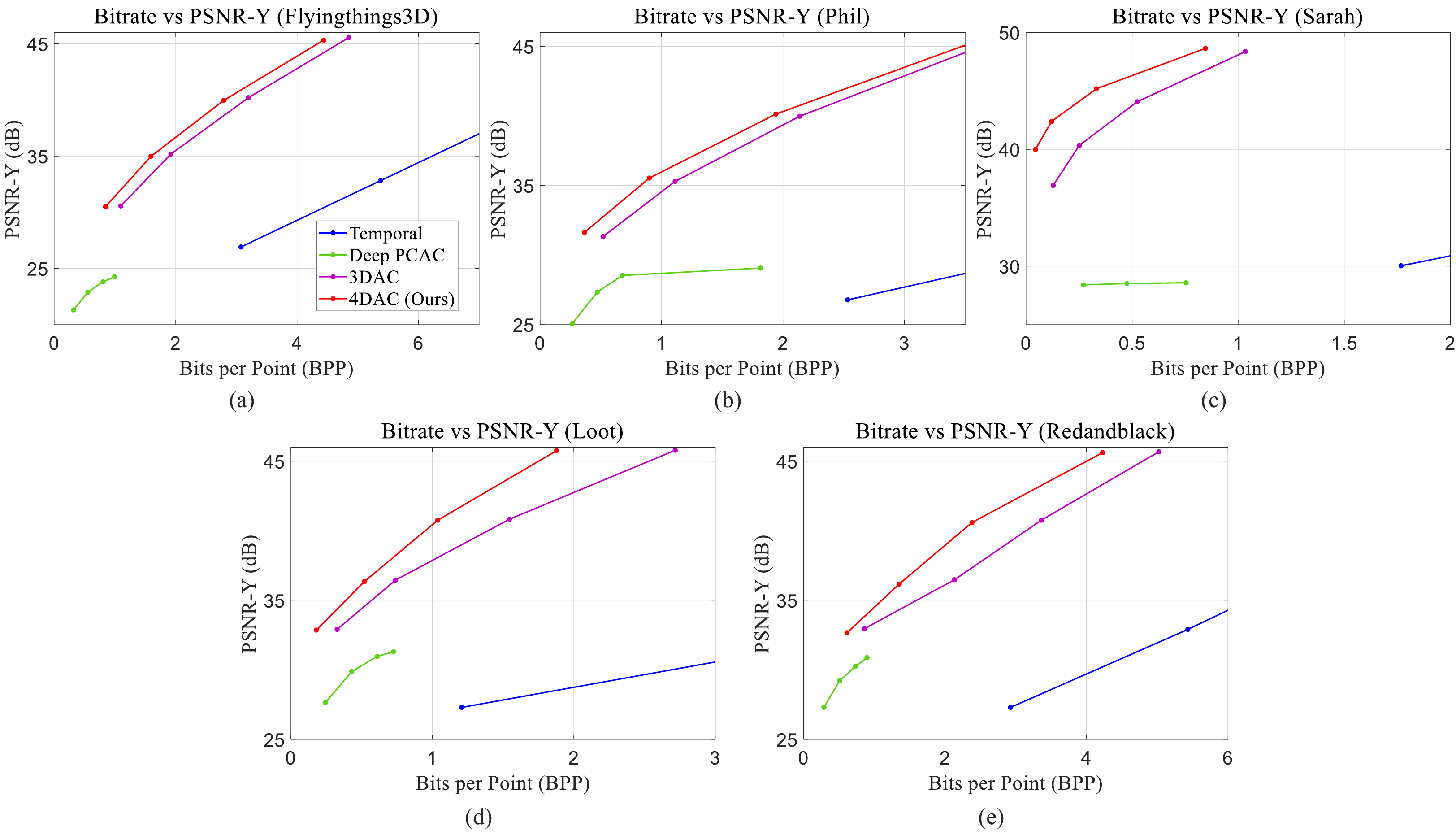}
    \caption{The quantitative results compared with learning-based methods. 
    }
    \label{fig:appendix_quant_dl}
\end{figure*}

We also compare our 4DAC with several learning-based attribute compression methods. Specifically, we additionally present the performance of DeepPCAC \cite{sheng2021deep} and Temporal. DeepPCAC is a static attribute compression method which constructs the transform coding block and the entropy model with point-based auto-encoders \cite{qi2017pointnet++}. Temporal is a simple dynamic attribute compression framework only containing a deep entropy model and an entropy coder. The entropy model extracts context feature from the previous point cloud through spatial-temporal aggregation inspired by \cite{biswas2020muscle}. As shown in Fig. \ref{fig:appendix_quant_dl}, our 4DAC significantly outperforms existing learning-based methods. This result demonstrates the effectiveness of our proposed method.


\subsection{Additional Qualitative Results}

\begin{figure*}[t]
	\centering
	\includegraphics[width=0.95\linewidth]{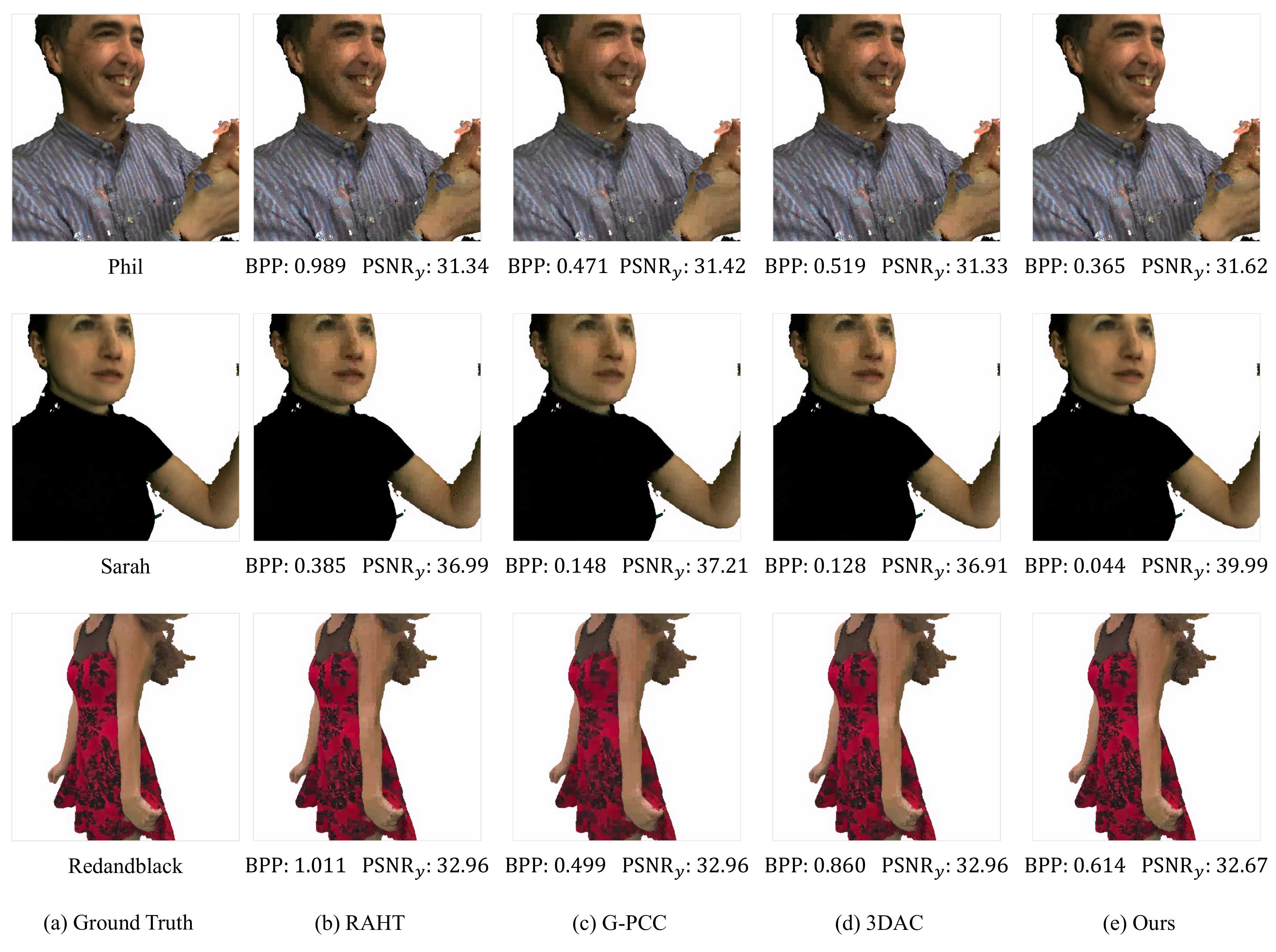}
    \caption{Additional qualitative results. 
    }
    \label{fig:appendix_qual}
\end{figure*}

In Fig. \ref{fig:appendix_qual}, We present additional qualitative results on MVUB "Phil", MVUB "Sarah" and 8iVFB "Redandblack". The corresponding $\rm BPP$ and $\rm {PSNR}_y$ of the sequence are also provided. It shows that our method can achieve better reconstruction quality with the help of the temporal information.


